\documentclass[conference,a4paper]{IEEEtran}
\usepackage{cite}
\usepackage{amsmath,amssymb,amsfonts}
\usepackage{graphicx}
\usepackage{booktabs}
\usepackage{tabularx}
\usepackage{array}
\usepackage[hyphens]{url}
\usepackage{tikz}
\usetikzlibrary{positioning,fit,backgrounds,arrows.meta,decorations.pathreplacing,calc}

\usepackage{dblfloatfix}

\begin{document}
\pagestyle{empty}

\title{Teaching Reinforcement Learning and Humanoid Robotics to High-School
Students: An Expert-Validated Curriculum
Design on a Low-Cost Open Platform}

\author{
\IEEEauthorblockN{Yuanzhe Dong}
\IEEEauthorblockA{Stanford University\\Stanford, CA, USA\\yzd@stanford.edu}
\and
\IEEEauthorblockN{Jie Cao}
\IEEEauthorblockA{University of North Carolina at Chapel Hill\\Chapel Hill, NC, USA\\jiecao@unc.edu}
\and
\IEEEauthorblockN{Shuman Wang}
\IEEEauthorblockA{Stanford University\\Stanford, CA, USA\\shuman@stanford.edu}
}

\maketitle
\thispagestyle{empty}

\begin{abstract}
Lower cost open source robots and reinforcement learning (RL) simulation tools
create new opportunities for precollege students to engage with contemporary
robotics. However, translating a complete research workflow, spanning
mechanical assembly, electrical setup, simulation, policy learning, system
identification, and physical deployment, into a coherent course for novice
learners remains challenging. We present an integrated robotics course
framework that organizes these activities around a shared robotic artifact.
The framework combines parallel disciplinary tracks, sequencing based on
technical dependencies, progressive integration of simulation and hardware,
layered performance checkpoints, and structures for balancing collaborative
work with individual accountability. We illustrate the framework through a
high school curriculum organized around a robot project in which pairs of
students assemble an open source humanoid robot, train a walking policy in
simulation, and deploy it on the physical platform. The framework was
developed through an iterative design process that included formative review
by five experts in robotics research, engineering, secondary STEM education,
and curriculum design. Expert feedback highlighted three central design
tensions: authenticity versus cognitive load, system integration versus timely
visible progress, and team construction versus individual accountability.
These tensions informed the final framework presented in this paper. This work
offers a structured approach for adapting robotics research workflows into
interdisciplinary precollege courses; future classroom studies are needed to
examine implementation and student learning.
\end{abstract}

\begin{IEEEkeywords}
Curriculum development, engineering education, humanoid robots,
pre-college engineering, reinforcement learning
\end{IEEEkeywords}

% =====================================================================
\section{Introduction}
% =====================================================================
\begin{figure}[!t]
\centering
\begin{minipage}{\columnwidth}
\centering
\begin{minipage}[b]{0.31\linewidth}
  \centering
  \includegraphics[width=\linewidth]{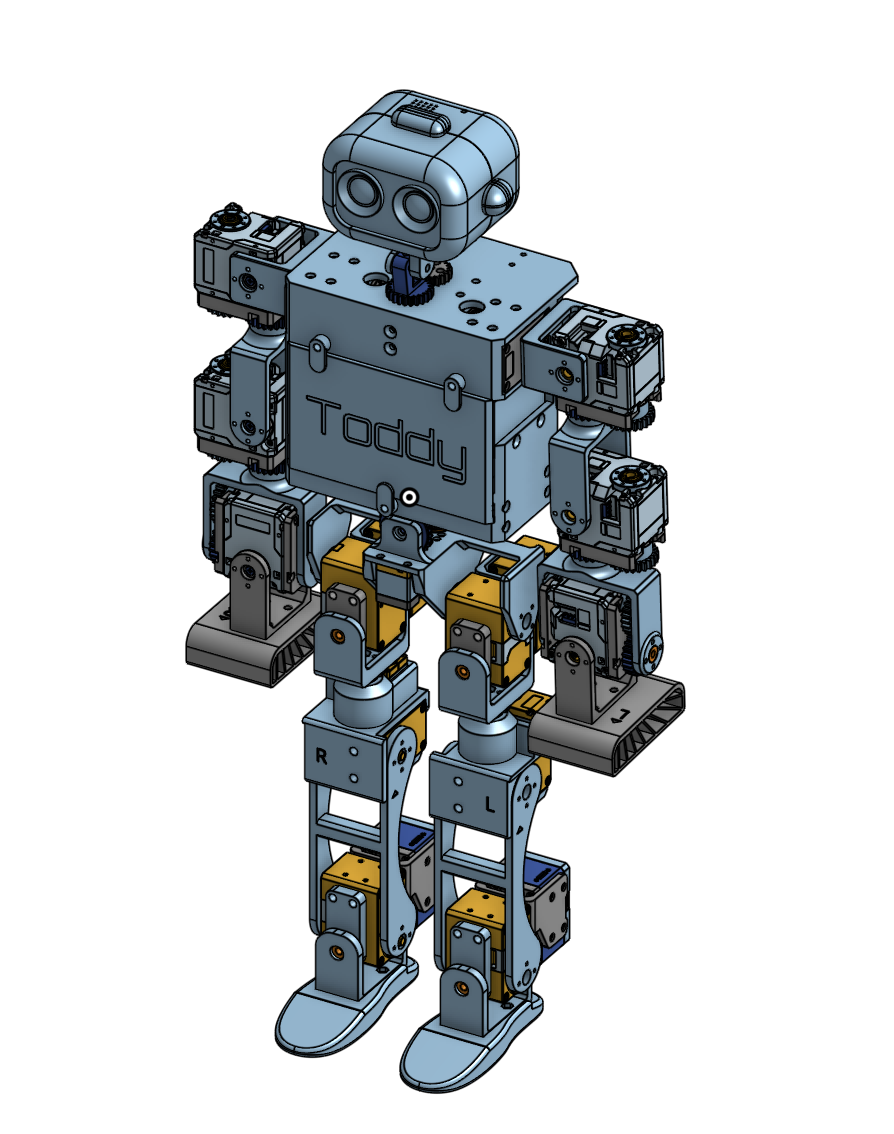}\\[2pt]
  {\footnotesize\makebox[\linewidth][c]{(a) Robot design}}
\end{minipage}\hfill
\begin{minipage}[b]{0.31\linewidth}
  \centering
  \includegraphics[width=\linewidth]{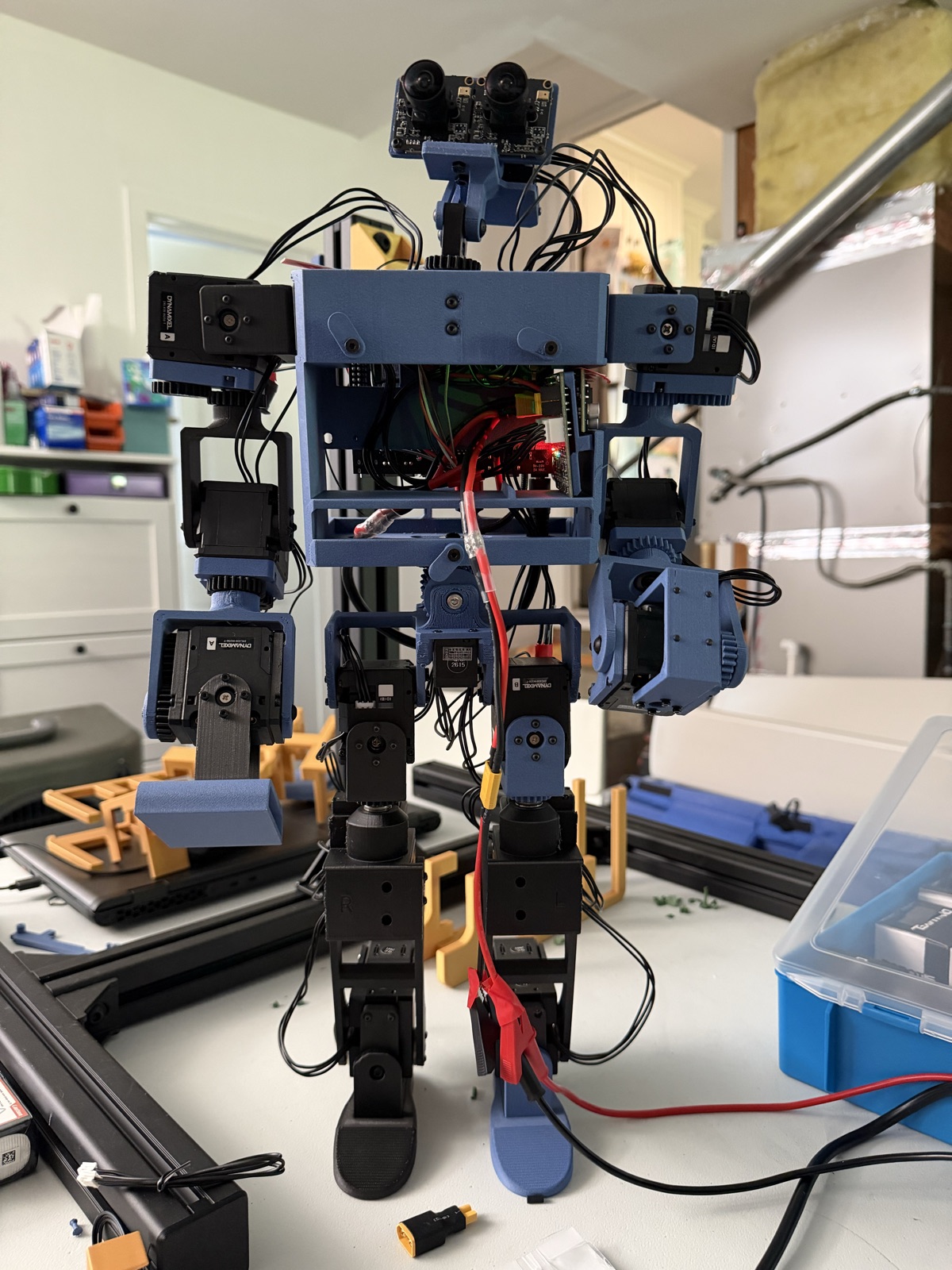}\\[2pt]
  {\footnotesize\makebox[\linewidth][c]{(b) Assembly in progress}}
\end{minipage}\hfill
\begin{minipage}[b]{0.31\linewidth}
  \centering
  \includegraphics[width=\linewidth]{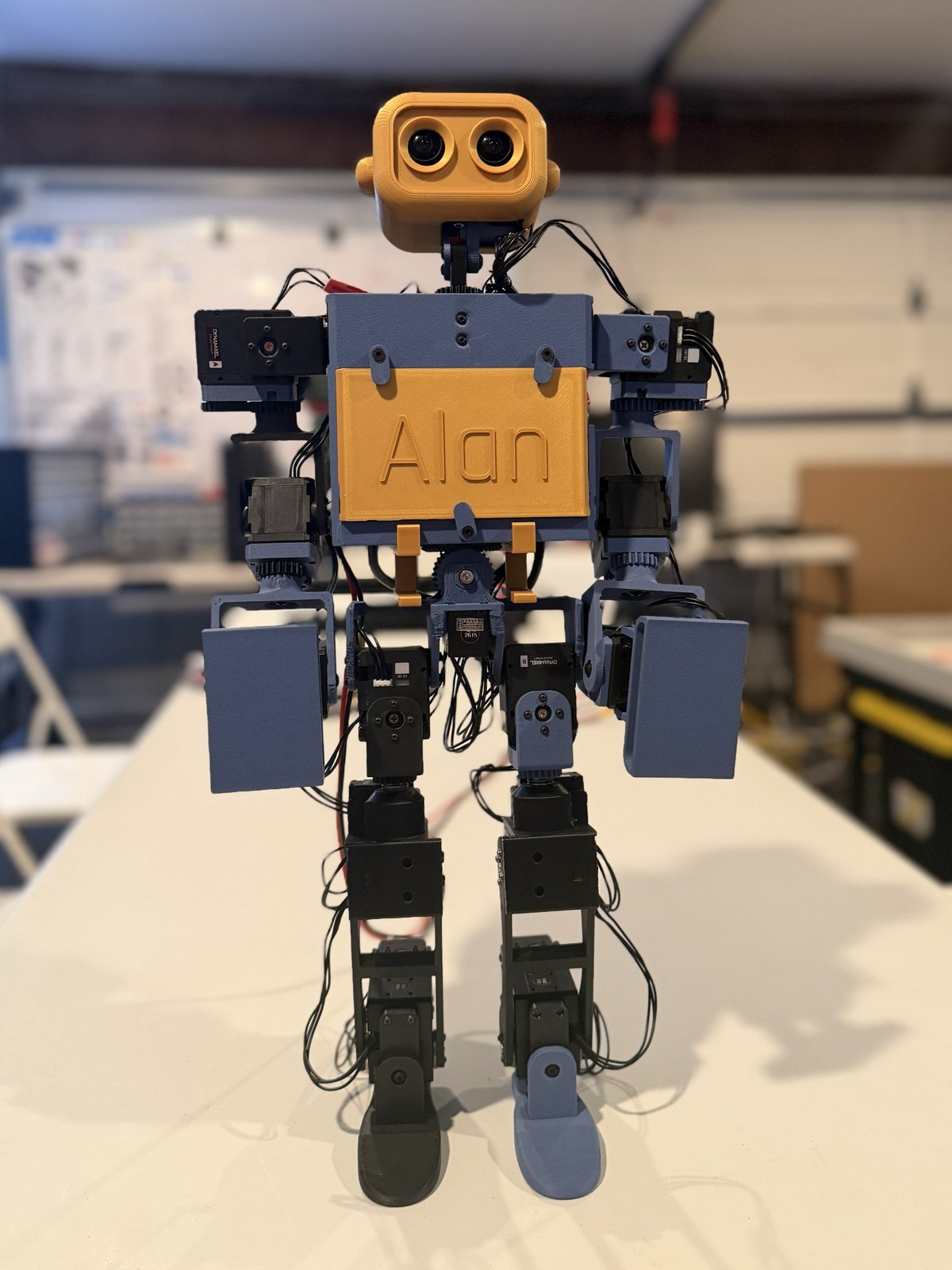}\\[2pt]
  {\footnotesize\makebox[\linewidth][c]{(c) Completed robot}}
\end{minipage}
\par\vspace{4pt}
\includegraphics[width=\linewidth]{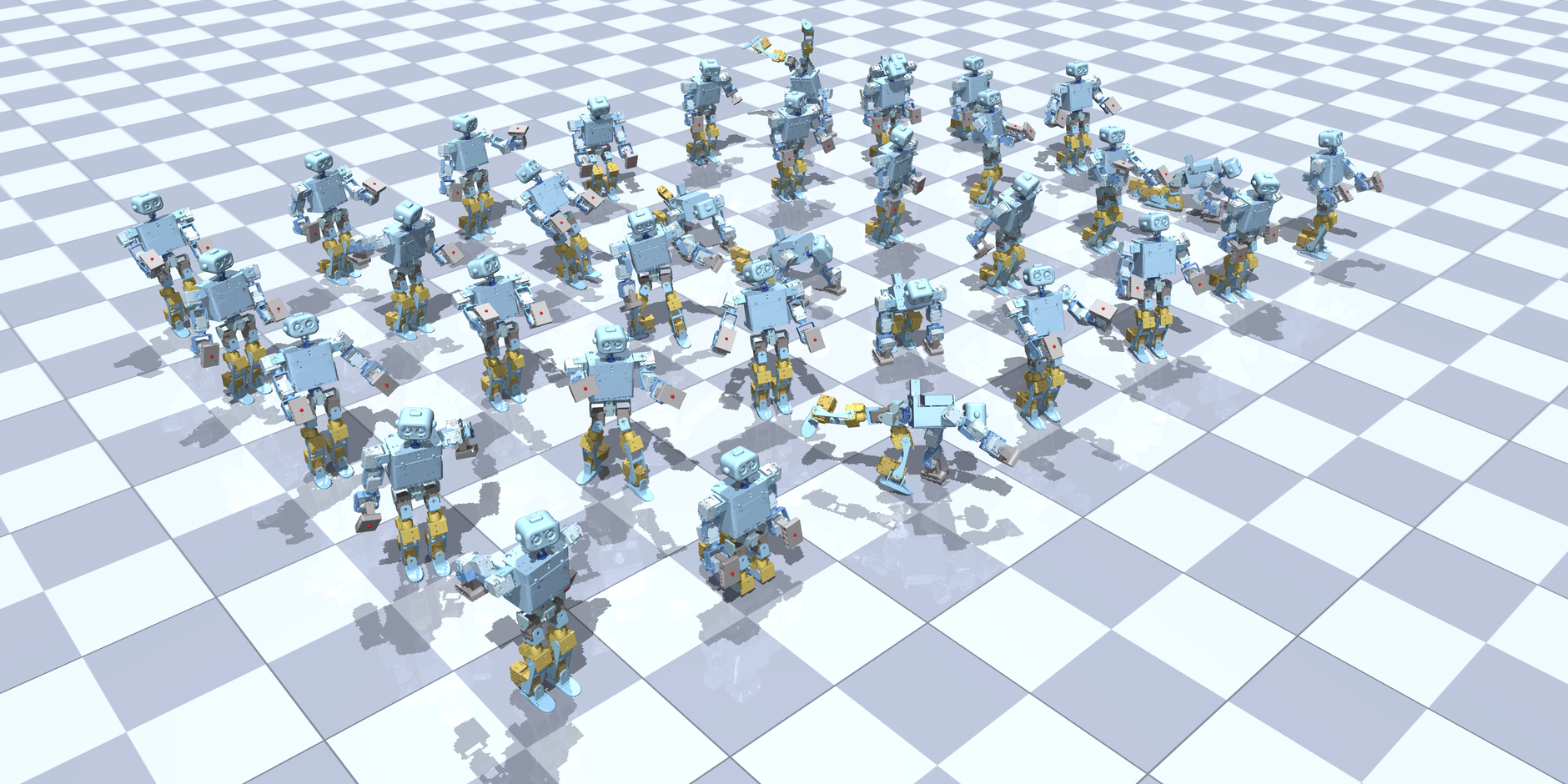}\\[2pt]
{\footnotesize (d) Learning in simulation}
\end{minipage}
\caption{Physical and virtual settings for the robot project: (a) robot
design, (b) assembly in progress, (c) completed robot, and (d) reinforcement learning in
simulation.}
\label{fig:platform}
\end{figure}

\IEEEPARstart{R}{obotics} connects programming and engineering, but teaching
reinforcement learning (RL) introduces hardware costs, space constraints,
and the distinction between programming and training
behavior~\cite{dietz2022,zhang2022}. Calls for deeper learner participation in
RL design~\cite{burman2025} motivate a framework connecting introductory
activities to a complete robot-learning workflow.

\emph{Why teach RL to high-school students?} Our focus on grades~9--12
serves three educational purposes. First, artificial intelligence (AI) literacy includes understanding
how systems acquire behavior and how human choices shape learning. The AI for K--12 (AI4K12) initiative's
draft guidance explicitly includes distinguishing supervised, unsupervised,
and reinforcement learning in this grade band~\cite{ai4k12learning}.
Second, RL activities can connect school mathematics and science to a
contemporary application: students interpret training curves, vary reward
weights, and test predictions about motion. Supplied code supports this
inquiry without requiring university-level optimization mathematics.
Third, defining and testing motion goals lets students explore computing and
engineering before further-study choices. Advanced algorithm implementation
is optional.

\emph{Why a framework now?} Open humanoids such as
ToddlerBot~\cite{toddlerbot2025} and Berkeley Humanoid Lite~\cite{berkeleylite2025}
combine hardware with tools for learning control strategies, or policies.
Industry tools such as NVIDIA Isaac Lab support simulation, training, and
testing~\cite{isaaclab}. Adapting these practices for high school requires
prerequisite support, manageable pacing, meaningful choices, and individual
assessment.

Our framework adapts the assembly-to-deployment workflow for high-school
students through staged tasks and bounded customization, making model
adjustment and physical testing explicit learning objectives. Contributions include
the framework, a humanoid curriculum instantiation, and a five-expert
formative review tracing feedback to revisions. Students completed the course
and achieved robot walking; their performance and learning data are reserved
for a separate study. This paper reports expert perspectives on the initial
design, not comparative learning gains.

% =====================================================================
\section{Related Work}
% =====================================================================
\emph{Accessible robotics and learner agency.} Pre-college robotics supports
construction and programming, but AI learning adds questions about how a
system acquires behavior. Torres et al.~\cite{torres2025} organize high-school
robotics around design-based learning. McLaughlin et al.~\cite{mclaughlin2026}
identify real-world practices, designing, and creative expression as pathways
in students' engagement with AI robotics. These approaches inform our use of
bounded choices alongside a common technical core; customization alone is
not our novelty.

\emph{Existing RL learning opportunities.} ARtonomous~\cite{dietz2022}
lets middle-school students train and customize virtual robot navigation,
avoiding physical kit costs. Zhang et al.~\cite{zhang2022} combine LEGO robots
and a web interface to teach RL to high-school students. Balancing
Act~\cite{burman2025} combines a physical LEGO task with simulated balance
control and lets learners explore reward and observation choices. These
studies establish that pre-college RL, physical robots, and customization
are already possible. Our additional design focus is following a policy
through a student-assembled humanoid's measurement, simulation training, and
physical deployment, with individual explanations at each stage.

\emph{Research workflows and the adaptation gap.} University courses provide
staged robotics labs~\cite{correll2013} and RL-based physical
experiments~\cite{podobnik2024}; open humanoids provide research control
pipelines~\cite{toddlerbot2025,berkeleylite2025}. Transferring these resources
to high school requires reducing prerequisite demands while preserving
meaningful decisions. The design problem is therefore access, coherence,
and assessable understanding across the workflow. Our framework extends the
scope of learning opportunities; it does not establish that a humanoid course
teaches introductory RL more effectively than simpler alternatives.

% =====================================================================
\section{Course Design Framework}
\label{sec:framework}
% =====================================================================
\subsection{Educational Rationale}
Integrated science, technology, engineering, and mathematics (STEM)
education connects disciplines within a meaningful context~\cite{kelley2016}.
The shared robot connects prediction, measurement, and explanation.
Scaffolding limits unfamiliar concepts and tools~\cite{sweller1998}, while
formative assessment uses explanations to guide instruction~\cite{black1998}.
These perspectives motivate the design; effectiveness remains untested.

\subsection{Design Principles}
The five principles below describe the framework. The three tensions in
Section~\ref{sec:revisions} organize expert feedback informing its revisions.

\emph{P1: Coordinate disciplines through one artifact.} Connect mechanical,
electrical, and computational work to observable behavior of the same robot.
For example, students explain a failed standing attempt using both balance
and control concepts. Teachers elicit these connections explicitly.

\emph{P2: Sequence by technical dependencies.} Make safety induction,
component checks, power-on, standing, and walking observable gates.
Simulation can proceed before hardware is ready, providing early conceptual
progress while assembly continues.

\emph{P3: Integrate simulation and hardware progressively.} Move from
simulated predictions to physical measurements. A supplied policy checks robot operation before a
student-trained policy is introduced, helping students distinguish equipment
faults, simulation errors, and learning failures.

\emph{P4: Combine scaffolded checkpoints with bounded choice.} Progress from
a worked example to a controlled modification and an independent decision.
Students interpret graphs, compare behavior, and justify choices of learning
objectives or non-critical design features within instructor-set limits.
Independent algorithm implementation is an extension.

\emph{P5: Pair collaboration with individual evidence.} Rotate operating
and checking roles so both learners predict, test, and explain. Record
individual reasoning separately from team robot performance, and use it to
adjust support before assigning a harder task.

\subsection{Intended Support for RL Understanding}
Learners investigate what rewards encourage, whether learned behavior
persists under changed conditions, and whether physical behavior matches
simulation. Comparing one controlled change with a baseline connects reward
design, generalization, and model limitations across virtual and physical
trials. These are proposed learning mechanisms, not measured gains.

% =====================================================================
\section{Humanoid Curriculum Instantiation}
% =====================================================================
\subsection{Implementation Model}
The curriculum targets grades~9--12, with four students working in two pairs,
one ToddlerBot per pair, and a recommended mentor-to-student ratio of 1:4.
An intake survey informs pairing and support. Initially, prior coding was
optional. Expert concerns prompted a revised prerequisite: running and
modifying a short Python program before RL labs. Beginners need an
introductory activity and additional preparation time.

Eight three-hour computer science and AI (CS+AI) sessions (24 contact hours)
accompany assembly, electrical checks, and a closing demonstration. Additional
build, preparation, and training time is not quantified here.

The intended learning outcomes are (LO1) assemble and safely start the
robot; (LO2) explain how rewards and trying different actions guide learning,
and distinguish training from using a learned policy; (LO3) compare simulation
with physical measurements and explain model adjustments and training under
varied conditions; (LO4) evaluate robot behavior and diagnose differences
between simulation and physical trials; and (LO5) collaborate and explain
design decisions. Independent algorithm implementation and successful
student-trained walking are extensions.

Integration serves these outcomes: diagnosing a walking failure (LO4)
connects assembly checks (LO1), reward choices (LO2), and simulation
comparisons (LO3). Pair work and individual explanations support LO5.
Table~\ref{tab:assessment} specifies evidence for each outcome.

\subsection{Platform}
ToddlerBot~\cite{toddlerbot2025} provides open hardware and software
(Fig.~\ref{fig:platform}), with a reported parts cost under USD~6{,}000.
Fabrication, repairs, computing, and mentoring add to delivery costs.
Shared computers support training. The platform's simulation, learning,
and measurement tools expose students to practices relevant to research
and industry~\cite{isaaclab}; professional proficiency is not claimed.

\emph{Proposed customization for high-school learners.} A progression of
choices offers creative and analytical entry points. Learners first select
colors or design non-critical shell features. Next, they define a bounded
motion objective, such as prioritizing steady walking over speed, modify
supplied reward weights, and compare simulation trials. An extension uses
an instructor-approved sequence of poses, such as a slow arm wave, with
rewards for following that sequence. The scaffold supplies the training code;
learners specify, test, and explain the behavior. New motions are evaluated
in simulation before any supervised hardware attempt. Motors and safety
settings remain fixed; shell changes must allow movement without excess weight.
These are proposed activities whose difficulty remains to be evaluated.

\subsection{Track Organization}
\begin{figure}[t]
\centering
\setlength{\tabcolsep}{1.2pt}%
\resizebox{\columnwidth}{!}{%
\begin{tabular}{@{}ccc@{}}
\includegraphics[width=0.30\columnwidth]{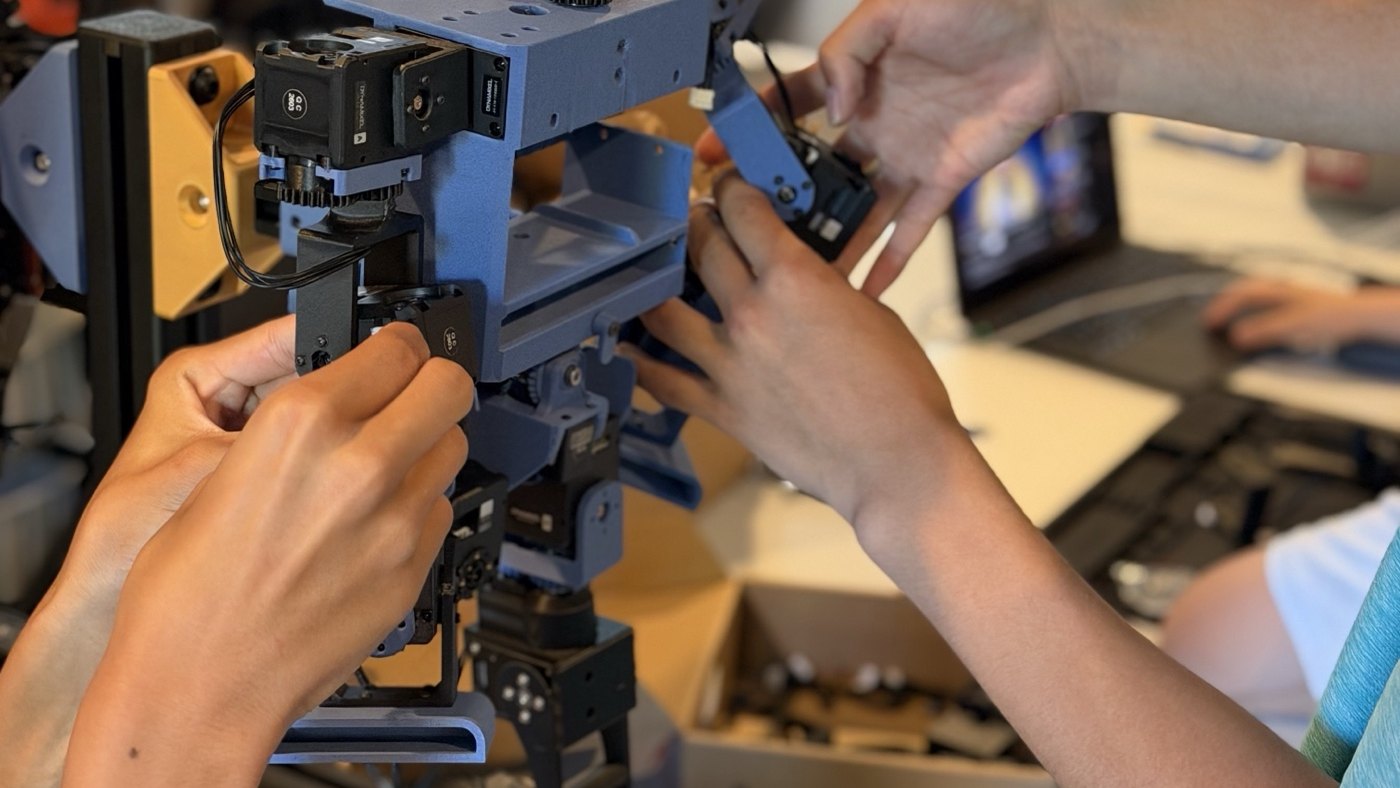} &
\includegraphics[width=0.30\columnwidth]{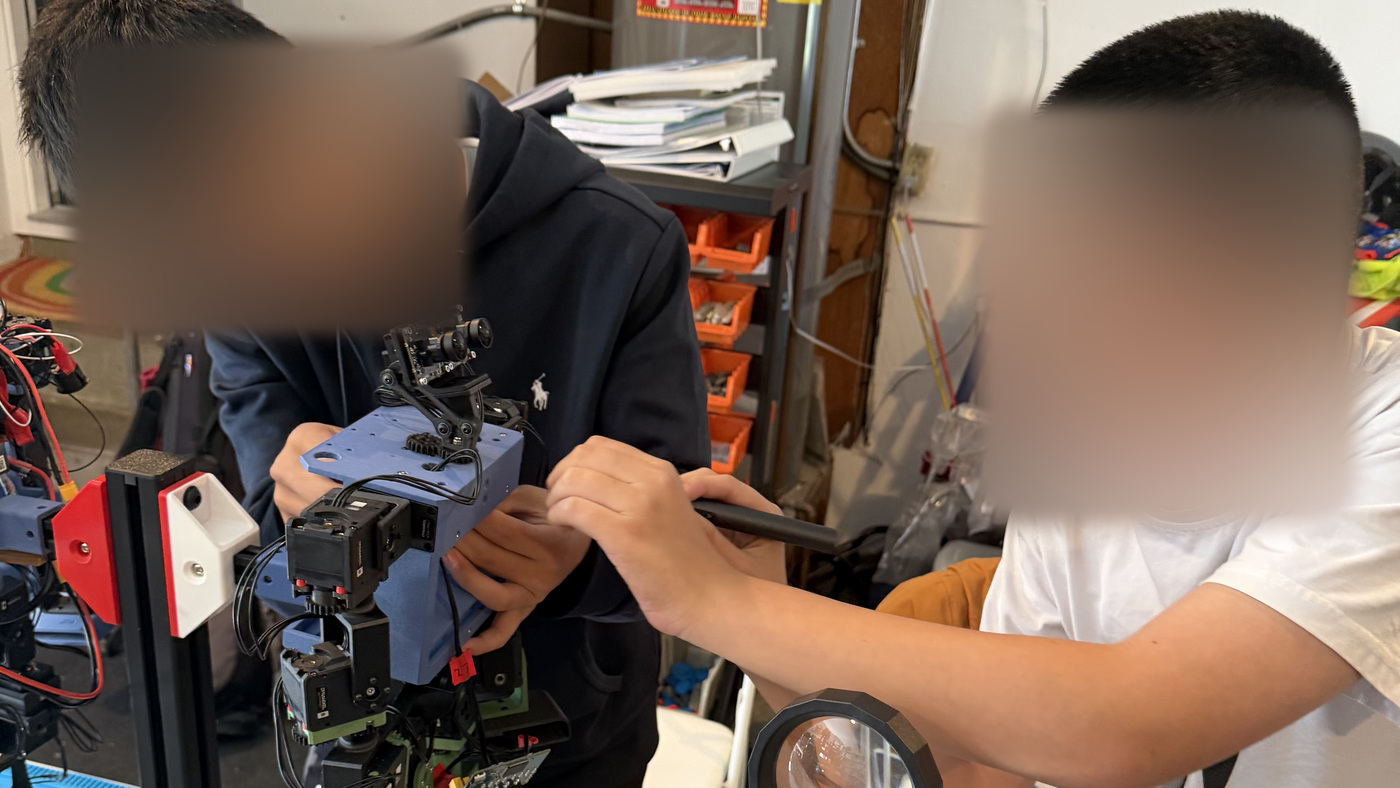} &
\includegraphics[width=0.30\columnwidth]{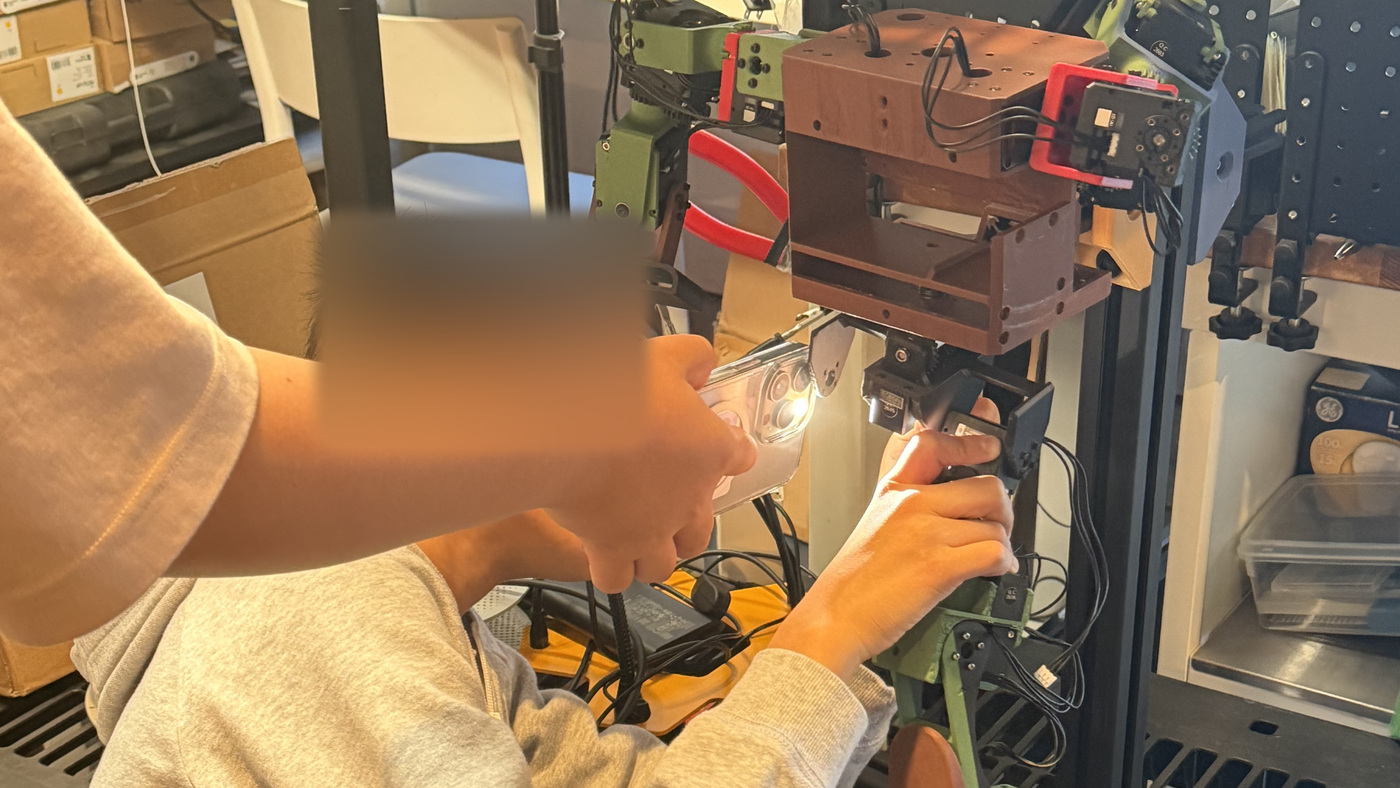} \\[1.2pt]
\includegraphics[width=0.30\columnwidth]{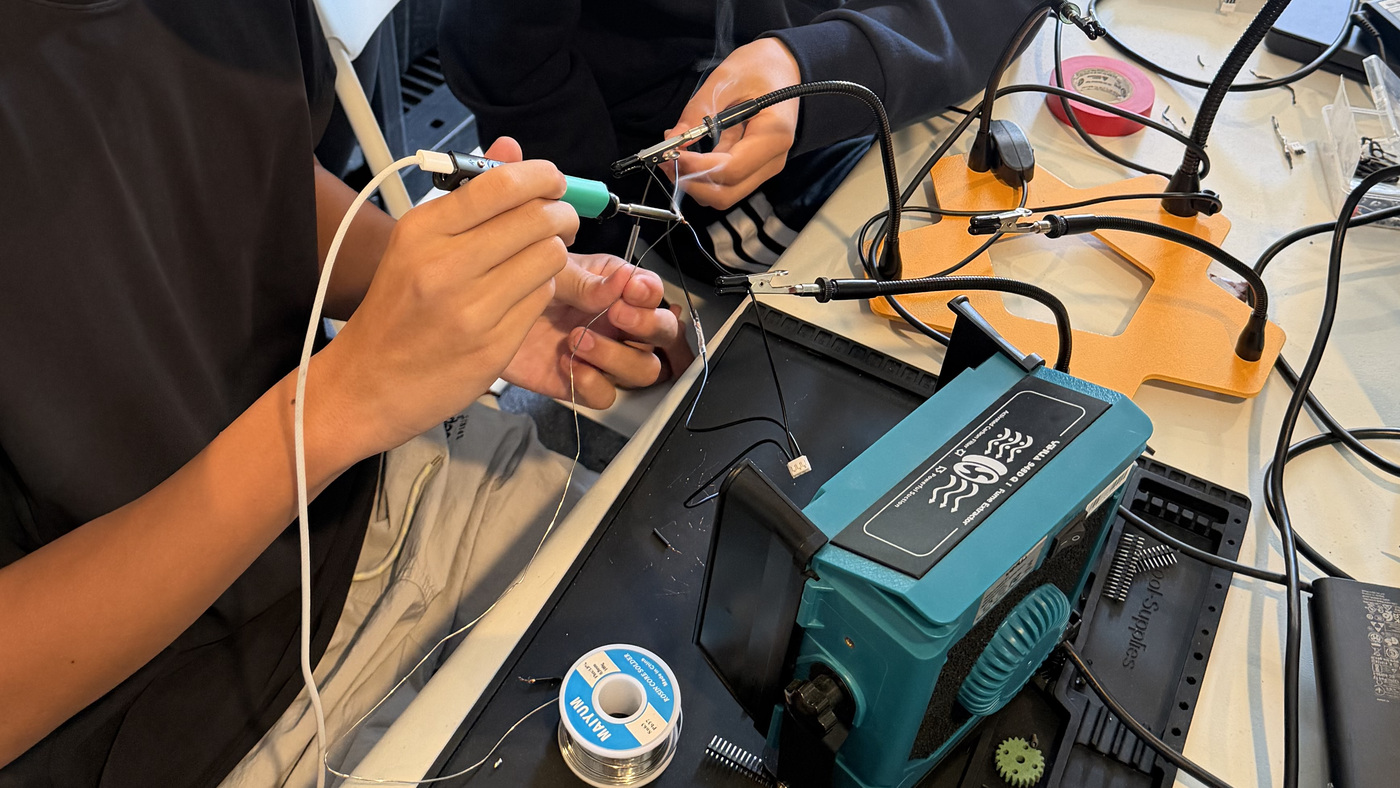} &
\includegraphics[width=0.30\columnwidth]{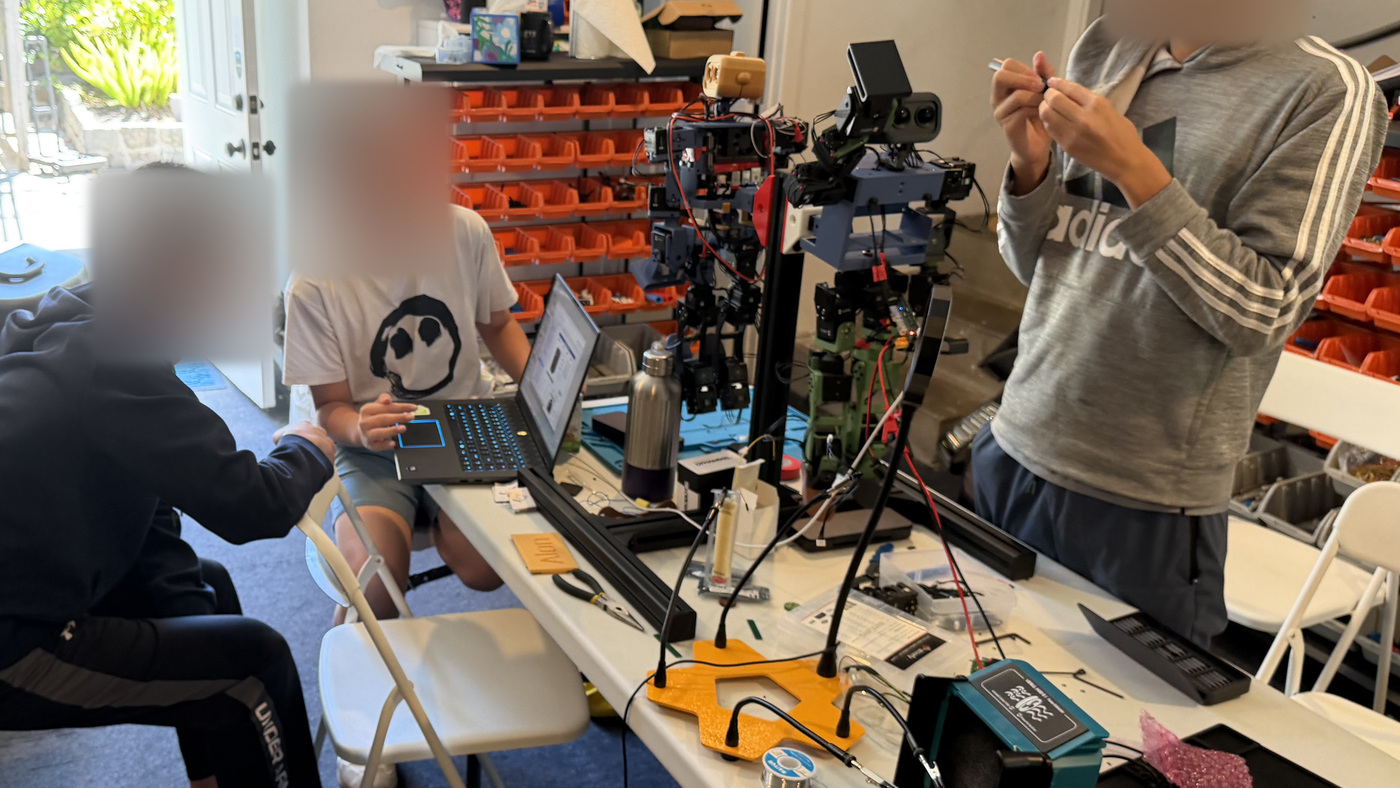} &
\includegraphics[width=0.30\columnwidth]{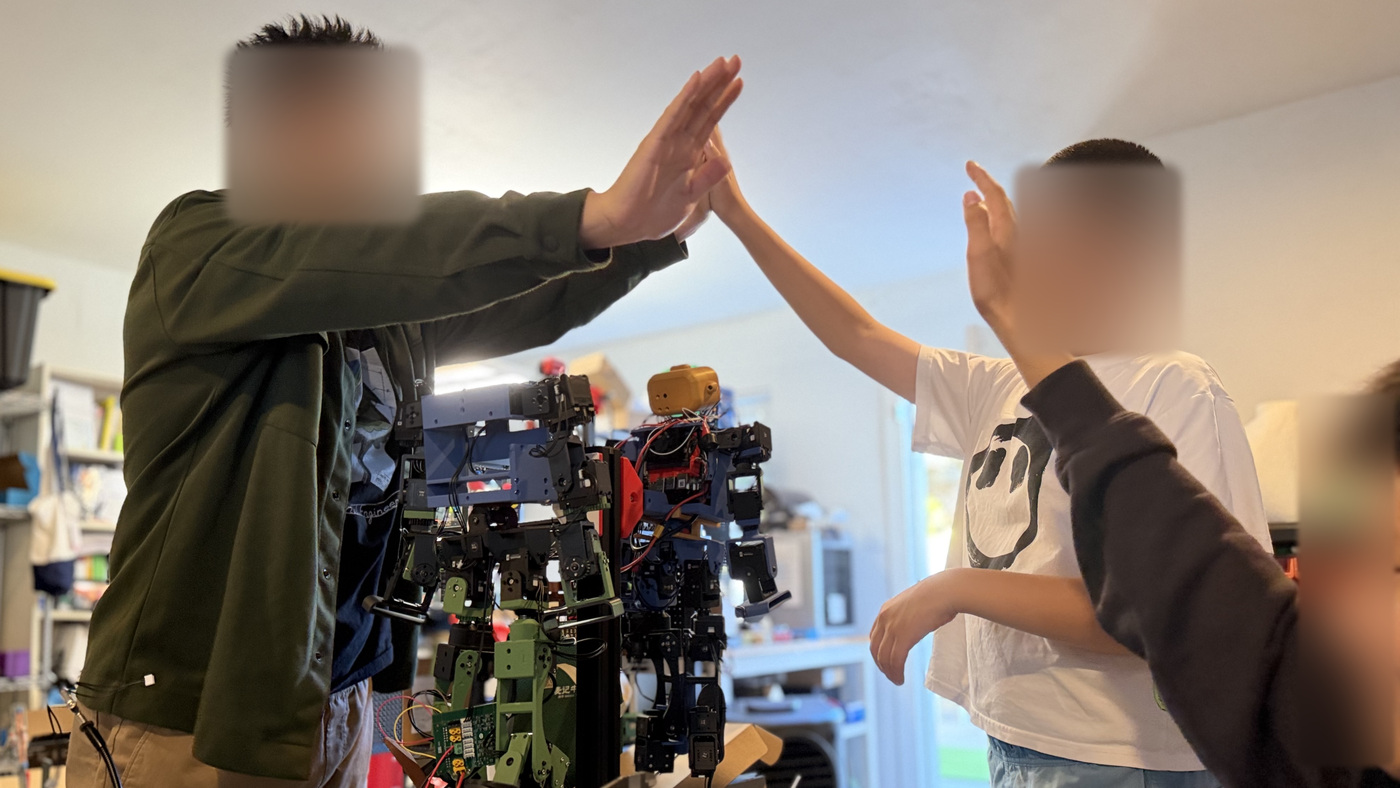}
\end{tabular}}
\caption{Students collaborating during hands-on assembly, soldering, and
on-robot testing, and celebrating with their teacher (bottom right).}
\label{fig:build}
\end{figure}

\begin{figure*}[t]
\centering
\resizebox{0.75\textwidth}{!}{%
\begin{tikzpicture}[
  font=\footnotesize,
  >={Stealth[length=2.4mm]},
  % box + label geometry is shared so every lane has identical visual weight
  box/.style={draw,rounded corners,minimum height=9mm,minimum width=23mm,
    text width=21mm,align=center,inner sep=2pt,line width=0.6pt},
  lbl/.style={draw,rounded corners,minimum height=9mm,minimum width=15mm,
    align=center,font=\footnotesize\bfseries,line width=0.6pt},
  % colour AND border-style differ per lane -> legible in grayscale / colourblind
  me/.style ={box,fill=blue!12,  draw=blue!60!black},
  meL/.style={lbl,fill=blue!22,  draw=blue!60!black},
  bu/.style ={box,fill=orange!25,draw=orange!80!black,dashed},
  buL/.style={lbl,fill=orange!38,draw=orange!80!black,dashed},
  ee/.style ={box,fill=violet!15,draw=violet!70!black,densely dotted},
  eeL/.style={lbl,fill=violet!26,draw=violet!70!black,densely dotted},
  cs/.style ={box,fill=teal!18,  draw=teal!75!black,very thick},
  csL/.style={lbl,fill=teal!30,  draw=teal!75!black,very thick},
  flow/.style={->,black!70,line width=0.6pt},
  merge/.style={black!55,line width=0.7pt},
  goal/.style={draw,rounded corners,minimum height=45mm,text width=24mm,
    align=center,fill=gray!15,draw=black,line width=1pt}
]
% ---- column x-centres (shared timeline) and row y-centres ----
\def\cA{2.6}\def\cB{5.1}\def\cC{7.6}\def\cD{10.1}
\def\yME{0}\def\yBU{-1.2}\def\yEE{-2.4}\def\yCS{-3.6}
% ---- faint column guides behind everything: "across a schedule" ----
\begin{scope}[on background layer]
  \foreach \x in {1.35,3.85,6.35,8.85,11.15}{
    \draw[gray!18,line width=0.4pt] (\x,0.7) -- (\x,-4.3);}
\end{scope}
% ---- top schedule axis (light cue that the lanes run in parallel over time) ----
\node[gray!65,font=\footnotesize\itshape] at (6.1,1.25) {conceptual progression};
\draw[->,gray!55,line width=0.6pt] (1.35,0.85) -- (11.15,0.85);
\node[gray!65,font=\scriptsize,left] at (1.35,0.85) {early};
\node[gray!65,font=\scriptsize,right] at (11.15,0.85) {late};
% ---- lane labels (common left edge) ----
\node[meL] (meL) at (0,\yME) {ME};
\node[buL] (buL) at (0,\yBU) {Build};
\node[eeL] (eeL) at (0,\yEE) {EE};
\node[csL] (csL) at (0,\yCS) {CS+AI};
% ---- ME: five theory topics, taught as openers, condensed to four stages ----
\node[me] (me1) at (\cA,\yME) {Design \& mechanisms};
\node[me] (me2) at (\cB,\yME) {Describing motion};
\node[me] (me3) at (\cC,\yME) {Explaining balance};
\node[me] (me4) at (\cD,\yME) {Motion \& control};
% ---- Build: official manual order ----
\node[bu] (bu1) at (\cA,\yBU) {Arm assembly};
\node[bu] (bu2) at (\cB,\yBU) {Arms \& legs};
\node[bu] (bu3) at (\cC,\yBU) {Head \& neck};
\node[bu] (bu4) at (\cD,\yBU) {Torso \& assembly checks};
% ---- EE: safety, bus/IDs, wiring, power-on ----
\node[ee] (ee1) at (\cA,\yEE) {Lab \& electrical safety};
\node[ee] (ee2) at (\cB,\yEE) {Connecting motors};
\node[ee] (ee3) at (\cC,\yEE) {Testing circuits};
\node[ee] (ee4) at (\cD,\yEE) {Safe power-on};
% ---- CS+AI: RL arc, condensed; bridge = single env -> parallel ----
\node[cs] (cs1) at (\cA,\yCS) {Virtual experiments};
\node[cs] (cs2) at (\cB,\yCS) {Learning from rewards};
\node[cs] (cs3) at (\cC,\yCS) {Training \& evaluation};
\node[cs] (cs4) at (\cD,\yCS) {Testing on the robot};
% ---- intra-lane progression arrows ----
\foreach \a/\b in {me1/me2,me2/me3,me3/me4}{\draw[flow](\a)--(\b);}
\foreach \a/\b in {bu1/bu2,bu2/bu3,bu3/bu4}{\draw[flow](\a)--(\b);}
\foreach \a/\b in {ee1/ee2,ee2/ee3,ee3/ee4}{\draw[flow](\a)--(\b);}
\foreach \a/\b in {cs1/cs2,cs2/cs3,cs3/cs4}{\draw[flow](\a)--(\b);}
% ---- convergence: merge through one junction, then a single arrow ----
\coordinate (J) at (11.9,-1.8);
\foreach \n in {me4,bu4,ee4,cs4}{\draw[merge] (\n.east) to[out=0,in=180] (J);}
\node[circle,fill=black,inner sep=1.4pt] at (J) {};
\node[goal] (robot) at (13.8,-1.8) {\textbf{Intended outcomes}\\[3pt]Working robot\\[3pt]Explain AI behavior\\Test predictions\\Justify decisions};
\draw[->,line width=1pt] (J) -- (robot.west);
\end{tikzpicture}}
\caption{Conceptual progression across Mechanical Engineering (ME), Build,
Electrical Engineering (EE), and Computer Science\,/\,AI (CS+AI), linking robot
performance with individual understanding. Safety checks constrain hardware
work; simulation can begin independently. This is not a synchronized timetable;
Table~\ref{tab:cs} details the eight CS+AI sessions.}
\label{fig:tracks}
\end{figure*}
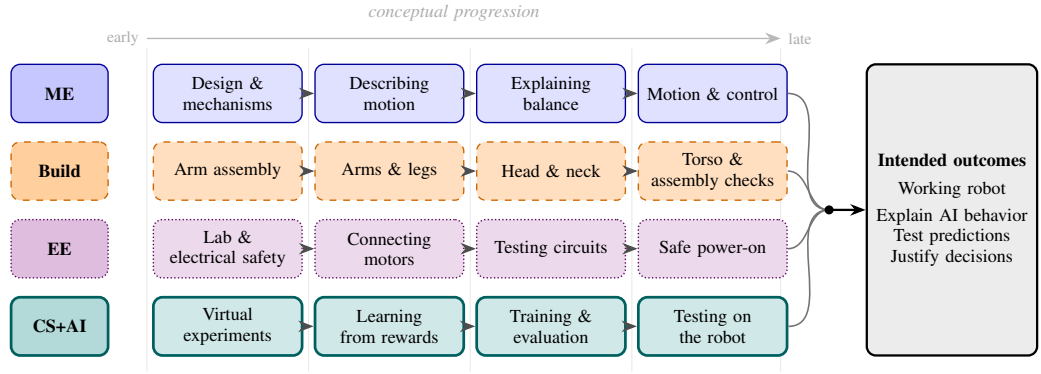

Four coordinated tracks connect disciplinary ideas to practical decisions
(Fig.~\ref{fig:tracks}). \emph{Mechanical Engineering (ME)} connects geometry,
balance, and motion to predictions about robot behavior. \emph{Build} turns
assembly and component checks into practice in interpreting instructions,
diagnosing faults, and documenting decisions (Fig.~\ref{fig:build}).
\emph{Electrical Engineering (EE)} develops safe working practices and
systematic testing before power-on. \emph{Computer Science/AI (CS+AI)} uses
simulation, learning, and physical comparison to examine how models and
rewards shape behavior.

Instructors introduce concepts immediately before their application and
return to them during debriefs. On-robot control requires the electrical
power-on gate, and walking requires a standing test. Simulation activities
can begin independently, providing early opportunities to predict and
interpret behavior while construction continues.

% =====================================================================
\subsection{CS+AI Activities and Checkpoints}
\label{sec:curriculum}
% =====================================================================
Table~\ref{tab:cs} connects eight sessions to guided activities and individual
explanations. Simulation (Fig.~\ref{fig:platform}(d)) supports repeated trials.
Students learn from demonstrated actions (imitation learning)
and rewards. Proximal policy optimization (PPO) updates a policy
using rewards~\cite{schulman2017}. Varying simulated
conditions (domain randomization) prepares policies for differences on the
physical robot~\cite{tobin2017,peng2018}. Students test a supplied policy before
their own; all physical trials require mentor supervision.

\begin{table*}[!t]
\caption{Revised Eight-Session CS+AI Plan (Proposed, Not Re-evaluated).
Provided code supports the core tasks; independent implementation and
student-trained walking are extensions.}

\label{tab:cs}
\centering
\scriptsize
\renewcommand{\arraystretch}{0.95}
\setlength{\tabcolsep}{4pt}
\begin{tabular}{p{0.4cm}p{3.0cm}p{6.8cm}p{6.8cm}}
\toprule
\# & Learning focus & Guided learning activity & Individual reflection / follow-up \\
\midrule
1 & Exploring learning in simulation &
Run a supplied simulation; identify what the robot senses, does, and receives
as a reward; predict the effect of changing a reward. &
Annotate the supplied code and explain the prediction. \\ \midrule
2 & Learning from examples &
Use supplied code to train from demonstrated actions (imitation learning);
compare the learned behavior with the examples. &
Report differences from the examples and explain one failed trial. \\ \midrule
3 & Learning from rewards &
Modify supplied training code (PPO); explain how rewards guide learning.
Extension: implement the training loop. &
Choose a bounded motion objective; interpret its learning curve. \\ \midrule
4 & Interpreting training progress &
Identify the same learning steps in software that runs many simulations
at once; launch a short run. Extension: adapt the reward. &
Use shared computers; save training graphs and trained versions. \\ \midrule
5 & Comparing simulation with reality &
After safety checks, measure motor behavior with the supplied tool; compare
physical measurements with simulation. &
Explain how adjusting a model setting changes agreement with measurements. \\ \midrule
6 & Testing changed conditions &
Use session-5 measurements to identify differences; change a simulation
setting and compare the learned behavior. &
Train under varied simulated conditions; predict a difficulty on the robot. \\ \midrule
7 & Testing a supplied behavior &
Run the supplied walking behavior; progress from supported checks to a
supervised walking attempt after the standing check. &
Record support conditions, trial duration, and a failure or adjustment. \\ \midrule
8 & Testing student choices &
Test the pair's trained behavior on the robot after the same checks; compare
with the supplied behavior. Extension: consecutive walking steps. &
Each student explains a motion-design choice and one observed discrepancy. \\
\bottomrule
\end{tabular}
\end{table*}

\emph{A proposed learning cycle.} Using supplied software, each learner
predicts the effect of a reward change. The pair trains and compares behavior
with the original. Each explains whether higher rewards mean better behavior;
the instructor uses these explanations to provide support or an extension.

Hardware targets are standing and approximately five seconds of walking with
the supplied policy, followed by an attempt with a student-trained policy.
Logs should distinguish supported stepping from unsupported walking; neither
is evidence of individual conceptual mastery. If hardware is unavailable,
simulation comparisons and measurement-log analysis preserve conceptual work
but do not substitute for the safe startup outcome (LO1).

Three-hour sessions alternate concept primers, supervised labs, and debriefs;
long training runs occur between sessions. Short physical trials and spare
parts accommodate damage from falls. Workload and reliability remain to be
evaluated.

% =====================================================================
\section{Formative Expert Review}
% =====================================================================
\subsection{Protocol and Analysis}
\begin{table}[t]
\caption{Expert Panel: Curriculum-Relevant Background}
\label{tab:experts}
\centering
\footnotesize
\begin{tabularx}{\columnwidth}{@{}l>{\raggedright\arraybackslash}X@{}}
\toprule
ID & Relevant background \\
\midrule
E1 & Master's in engineering; mechanical engineering, robotics,
industrial systems, and cross-disciplinary integration. \\ \midrule
E2 & Current full-time STEM teacher; master's in engineering;
secondary mathematics and science teaching. \\ \midrule
E3 & Current STEM teacher; master's degrees in computer science and
education; high-school teaching, STEM curriculum evaluation,
fabrication, and digital-engineering projects. \\ \midrule
E4 & Doctoral-level robotics research; open humanoid development,
simulation, and transferring learned behavior to physical robots. \\ \midrule
E5 & Doctoral-level computer-science research; electrical engineering,
robot perception, RL, and university teaching assistance. \\
\bottomrule
\end{tabularx}
\end{table}

\begin{table*}[!t]
\caption{Design Before and After Expert Review. Revised Responses Are Proposed
and Have Not Been Re-evaluated. T1--T6 Refer to the Qualitative Themes.}
\label{tab:revisions}
\centering
\footnotesize
\setlength{\tabcolsep}{4pt}
\renewcommand{\arraystretch}{1.08}
\begin{tabularx}{\textwidth}{@{}p{2.1cm}>{\raggedright\arraybackslash}X>{\raggedright\arraybackslash}X>{\raggedright\arraybackslash}X@{}}
\toprule
Design tension & Initial curriculum & Expert feedback & Revised design response \\
\midrule
Authenticity vs.\ cognitive load &
Prior coding helpful but optional; students implement PPO and adapt it to software
running many simulations at once within the first four CS+AI sessions. &
The integrated workflow was valued (T1), but four experts questioned novice
workload and prerequisites (T2); setup support was needed (T3). &
Retain authentic tasks with supplied code, preparation, and bounded reward
or shell customization. Make independent implementation an extension;
assess explanations before increasing complexity (P1, P4). \\ \midrule
Integration vs.\ visible progress &
Assembly and startup checks lead to physical testing, with a pretrained policy as
an intermediate checkpoint. Early conceptual evidence is less explicit. &
Setup can delay visible success (T2); reliable resources and backup activities
are needed (T3). Three experts requested clearer relevance and progress (T6). &
Retain hardware gates and the pretrained checkpoint. Add explicit early
simulation predictions and component comparisons; use standard configurations
and backup conceptual tasks (P2, P3). \\ \midrule
Team construction vs.\ individual accountability &
Pairs share a robot, with performance gates and technical logs; individual
roles and assessment criteria are underspecified. &
Unequal participation and opaque individual reasoning concerned experts
(T4--T5). Communication and assessment averaged 3.40 and 3.60/5. &
Rotate operating/checking roles and record contributions. Require individual
predictions, failure explanations, and design justifications; assess these
separately from team robot performance (P5). \\
\bottomrule
\end{tabularx}
\end{table*}

Five experts in robotics research and engineering, secondary STEM teaching,
and curriculum design evaluated the curriculum through structured interviews.
We report responses and backgrounds using codes E1--E5
(Table~\ref{tab:experts}). Experts are acknowledged by name with permission,
without linking names to codes.

The ratings and qualitative findings concern the initial curriculum, before
the scaffolding, outcome, and assessment refinements presented here. Experts
did not evaluate the revised session plan or the framework as a separate
instrument. The three tensions below are an author synthesis of the findings.

Each expert reviewed the initial four-track structure, eight-session CS+AI
plan, safety checkpoints, and staged deliverables. The protocol was adapted from the Integrated STEM Curriculum Planning and
Reflection Rubric~\cite{walker2018}, which derives from earlier engineering
design-based STEM curriculum assessment work~\cite{guzey2016}. It elicited
ratings on eight dimensions: (1)~motivating and engaging context,
(2)~integration of AI/robotics content, (3)~instructional strategies,
(4)~teamwork, (5)~communication of technical reasoning, (6)~formative and
summative assessment, (7)~curriculum organization, and (8)~integration of
educational technology. All dimensions used the same five-point scale (1 = not
adequate; 5 = excellent). Open-ended prompts then asked experts to explain each
rating and identify strengths, implementation conditions, and revision
priorities. An additional item asked for an overall implementation-readiness
rating. We summarized ratings descriptively and coded substantive responses
iteratively into cross-case themes. One expert omitted the overall-readiness
rating (no value was imputed), and a duplicated interview record was counted
only once.

\subsection{Quantitative Findings}
\begin{figure}[t]
  \centering
  \includegraphics[width=\columnwidth]{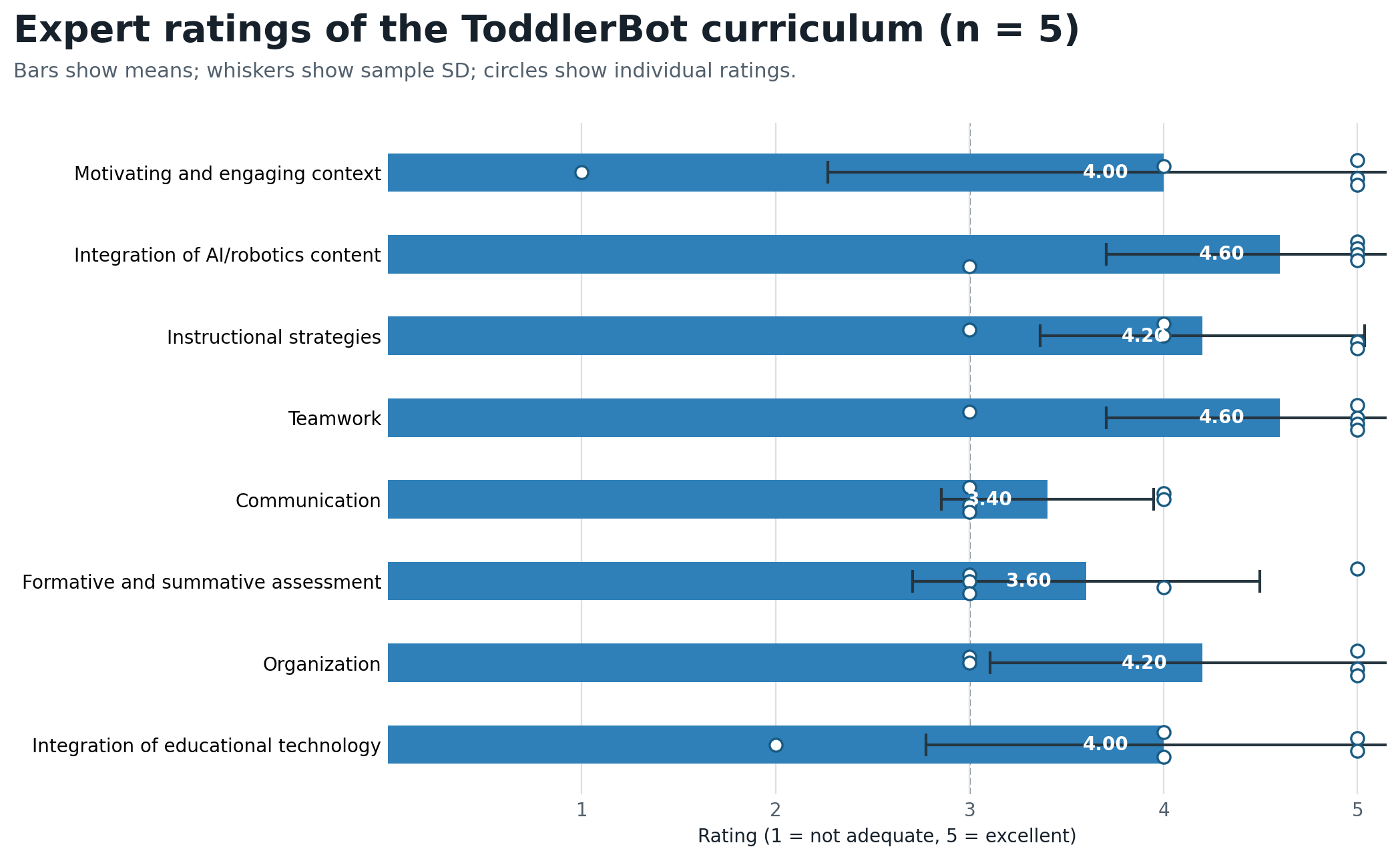}
  \caption{Expert ratings of the initial design across eight dimensions ($n=5$). Bars show means ($M$),
  whiskers show sample standard deviations ($SD$), and circles show individual
  ratings on a five-point scale.}
  \label{fig:expert-ratings}
\end{figure}

Experts rated the initial design favorably on integration and teamwork
(Fig.~\ref{fig:expert-ratings}). Integration of AI/robotics content and teamwork
were the highest-rated dimensions (both $M=4.60$, $SD=0.89$), followed by
instructional strategies and organization (both $M=4.20$). Motivating context
and educational-technology integration each averaged 4.00; motivating context
showed the greatest dispersion ($SD=1.73$, range 1--5).
Assessment ($M=3.60$, $SD=0.89$) and communication ($M=3.40$, $SD=0.55$) were
comparatively weaker. Among the four experts who supplied an overall score,
implementation readiness averaged 4.00 ($SD=0.82$, range 3--5).

\subsection{Qualitative Findings}
Six cross-case themes summarize the feedback. Counts below indicate the
number of experts expressing a theme, not its measured importance.

\emph{T1: Coherence.} All five valued connecting assembly, electronics,
simulation, and deployment through one artifact, while favoring a smaller
conceptual core over equal depth in every discipline.

\emph{T2: Cognitive load and delayed progress.} Four questioned the volume
of terminology and prerequisites, long sessions, and setup work that could
delay visible success. The motivation rating of 1/5 cautions against treating
the favorable mean as consensus about engagement.

\emph{T3: Resource demands.} All five identified implementation conditions,
including reliability, setup, expertise, supervision, or cost. Suggestions
included validated instructions, configured software, and backup activities.

\emph{T4: Unequal participation.} Experts supported pair work but warned
that differences in experience could concentrate technical work in one
student. Formation of complementary pairs and rotating roles were suggested.
One expert could not confidently judge teamwork because individual
responsibility was underspecified.

\emph{T5: Reasoning beyond completion.} Experts distinguished passing a
robot milestone from understanding it. They wanted explanations of failures
and design decisions, with prompts that elicit interpretation of logs and
measurement results.

\emph{T6: Visible relevance.} Three recommended connecting concepts to
observable behavior and future study opportunities, using demonstrations
and frequent achievements to sustain engagement.

\subsection{Design Tensions and Framework Refinements}
\label{sec:revisions}
Table~\ref{tab:revisions} connects expert feedback to revisions through three
tensions synthesized from the six themes. These tensions are author
interpretations; the comparison documents design changes, not learning gains.

\emph{Proposed individual assessment during pair work.} Students alternate
operating and checking roles and record their contributions. At each
checkpoint, each submits a separate prediction and explanation of the results.
The instructor observes each student's relevant practical task and asks a
brief follow-up question without partner assistance. Using
Table~\ref{tab:assessment}, the instructor records each outcome as independently
demonstrated, demonstrated with prompting, or not yet demonstrated, with
supporting evidence. Role logs inform assessment of collaboration but do not
establish understanding. Pair robot performance is recorded separately;
partners need not receive the same individual judgment. Students needing
support receive feedback and repeat the relevant task or explanation.
This assessment procedure is proposed and remains to be evaluated.

\begin{table}[t]
\caption{Proposed Individual Evidence and Success Criteria}
\label{tab:assessment}
\centering
\footnotesize
\begin{tabularx}{\columnwidth}{@{}l>{\raggedright\arraybackslash}X@{}}
\toprule
LO & Evidence and criterion \\
\midrule
1 & Observed safety task: follows the checklist and explains
when power-on must be withheld. \\ \midrule
2 & Annotated reward experiment: explains exploration and reward effects,
and distinguishes training from policy execution. \\ \midrule
3 & Simulation comparison: identifies a measured difference and explains
a model adjustment and why training conditions are varied. \\ \midrule
4 & Trial analysis: uses observations to support a failure
hypothesis and proposes a test that could distinguish causes. \\ \midrule
5 & Role log and individual explanation: documents contributions and
communicates a design decision to a non-specialist. \\
\bottomrule
\end{tabularx}
\end{table}

% =====================================================================
\section{Discussion and Conclusion}
% =====================================================================
\emph{Limits of the evidence.} Five purposively selected experts provide
formative feedback, not representative agreement. Platform expertise may
favor the approach; independence should not be assumed. Ratings establish
neither formal content validity nor learning effectiveness. Course completion
and walking provide implementation context; student outcomes are reserved
for a separate study. The revised framework and assessment criteria remain
unevaluated.

\emph{Limits of transfer.} Other platforms require adapted prerequisites,
checkpoints, and assessments. Mentoring, training compute, and repair capacity
remain resource assumptions; classroom studies must measure workload and
success rates.

Comparisons with simulation-only RL should account for prior experience,
contact time, and instructor support. Shared tasks should assess reward design,
training versus evaluation, unfamiliar failures, completion time, assistance,
and student explanations. Any learning advantage remains untested.

\section*{Acknowledgment}
Following peer review, GPT-6 Astra~\cite{gpt6astra} and Claude Fable
5.1~\cite{claudefable51} assisted with drafting
and revising the abstract and Sections I--VI, improving language and clarity,
and shortening text to meet the page limit. The authors reviewed and revised
all AI-assisted text for accuracy and take responsibility for the final manuscript.

The authors thank Haochen Shi and Weizhuo Wang for ToddlerBot technical
support, and Haotong Han, Yao He, Yi Yang, Miaoya Zhong, and Haochen Shi for
their interview participation and feedback on the curriculum.

% =====================================================================


\begin{thebibliography}{00}

\bibitem{dietz2022}
G. Dietz, J. King Chen, J. Beason, M. Tarrow, A. Hilliard, and R. B. Shapiro,
``ARtonomous: Introducing middle school students to reinforcement learning
through virtual robotics,'' in \textit{Proc. 21st Annu. ACM Interaction Design
and Children Conf. (IDC)}, Braga, Portugal, 2022, doi:~10.1145/3501712.3529736.

\bibitem{zhang2022}
Z. Zhang, S. Willner-Giwerc, J. Sinapov, J. Cross, and C. Rogers,
``An interactive robot platform for introducing reinforcement learning to
K-12 students,'' in \textit{Robotics in Education}, AISC, vol.~1359,
2022, pp.~288--301, doi:~10.1007/978-3-030-82544-7\_27.

\bibitem{burman2025}
T. Burman, J. Coney, C. Rogers, J. Cross, and J. Sinapov,
``Balancing act: Mastering beam-and-ball control with reinforcement learning,''
in \textit{Robotics in Education}, LNNS, vol.~1544, 2025, pp.~240--252,
doi:~10.1007/978-3-031-98762-5\_21.

\bibitem{ai4k12learning}
AI4K12 Initiative, ``Big idea 3: Learning,'' draft grade-band progression
chart. [Online]. Available: \url{https://ai4k12.org/big-idea-3-overview/}.
Accessed: Sep.~18, 2026.

\bibitem{toddlerbot2025}
H. Shi, W. Wang, S. Song, and C. K. Liu, ``ToddlerBot: Open-source
ML-compatible humanoid platform for loco-manipulation,'' in \textit{Proc.
9th Conf. Robot Learning (CoRL)}, PMLR, vol.~305, 2025, pp.~4165--4189.

\bibitem{berkeleylite2025}
Y. Chi, Q. Liao, J. Long, X. Huang, S. Shao, B. Nikoli\'c, Z. Li, and
K. Sreenath, ``Demonstrating Berkeley Humanoid Lite: An open-source,
accessible, and customizable 3D-printed humanoid robot,'' in \textit{Proc.
Robotics: Science and Systems (RSS)}, 2025, doi:~10.15607/RSS.2025.XXI.062.

\bibitem{isaaclab}
NVIDIA, ``Isaac Lab,'' [Online]. Available:
\url{https://developer.nvidia.com/isaac/lab}. Accessed: Sep.~18, 2026.

\bibitem{torres2025}
A. Torres \textit{et al.}, ``Teaching collaborative robotics: Design and
evaluation of design-based learning curriculum for high school STEM
education,'' \textit{J. Formative Design Learn.}, vol.~9, pp.~77--91, 2025.

\bibitem{mclaughlin2026}
G. McLaughlin \textit{et al.}, ``Pathways to learning: Exploring high school
students' learning of AI-powered educational robotics,''
\textit{Educ. Technol. Res. Dev.}, 2026,
doi:~10.1007/s11423-026-10651-w.

\bibitem{correll2013}
N. Correll, R. Wing, and D. Coleman, ``A one-year introductory robotics
curriculum for computer science upperclassmen,'' \textit{IEEE Trans. Educ.},
vol.~56, no.~1, pp.~54--60, Feb. 2013.

\bibitem{podobnik2024}
J. Podobnik, A. Udir, M. Munih, and M. Mihelj, ``Teaching approach for deep
reinforcement learning of robotic strategies,'' \textit{Comput. Appl. Eng.
Educ.}, vol.~32, no.~6, Art.~no.~e22780, 2024.

\bibitem{kelley2016}
T. R. Kelley and J. G. Knowles, ``A conceptual framework for integrated STEM
education,'' \textit{Int. J. STEM Educ.}, vol.~3, no.~1, Art.~no.~11, 2016.

\bibitem{sweller1998}
J. Sweller, J. J. G. van Merri\"enboer, and F. G. W. C. Paas, ``Cognitive
architecture and instructional design,'' \textit{Educ. Psychol. Rev.},
vol.~10, no.~3, pp.~251--296, 1998.

\bibitem{black1998}
P. Black and D. Wiliam, ``Assessment and classroom learning,''
\textit{Assess. Educ.: Princ., Policy \& Pract.}, vol.~5, no.~1, pp.~7--74,
1998.

\bibitem{schulman2017}
J. Schulman, F. Wolski, P. Dhariwal, A. Radford, and O. Klimov, ``Proximal
policy optimization algorithms,'' \textit{arXiv:1707.06347}, 2017.

\bibitem{tobin2017}
J. Tobin, R. Fong, A. Ray, J. Schneider, W. Zaremba, and P. Abbeel, ``Domain
randomization for transferring deep neural networks from simulation to the real
world,'' in \textit{Proc. IEEE/RSJ Int. Conf. Intell. Robots Syst. (IROS)},
2017, pp.~23--30.

\bibitem{peng2018}
X. B. Peng, M. Andrychowicz, W. Zaremba, and P. Abbeel, ``Sim-to-real transfer
of robotic control with dynamics randomization,'' in \textit{Proc. IEEE Int.
Conf. Robot. Autom. (ICRA)}, 2018, pp.~3803--3810.

\bibitem{walker2018}
W. S. Walker, III, T. J. Moore, S. S. Guzey, and B. H. Sorge, ``Frameworks to
develop integrated STEM curricula,'' \textit{K-12 STEM Education}, vol.~4,
no.~2, pp.~331--339, 2018.

\bibitem{guzey2016}
S. S. Guzey, T. J. Moore, and M. Harwell, ``Building up STEM: An analysis of
teacher-developed engineering design-based STEM integration curricular
materials,'' \textit{J. Pre-College Eng. Educ. Res.}, vol.~6, no.~1,
Art.~no.~2, 2016.

\bibitem{gpt6astra}
OpenAI, ``GPT-6 Astra.'' [Online]. Available:
\url{https://developers.openai.com/api/docs/models/gpt-6-astra}.
Accessed: Sep.~19, 2026.

\bibitem{claudefable51}
Anthropic, ``Claude Fable 5.1.'' [Online]. Available:
\url{https://platform.claude.com/docs/en/models/fable-5-1/overview}.
Accessed: Sep.~19, 2026.

\end{thebibliography}
\end{document}